\documentclass[letterpaper, 10 pt, conference]{ieeeconf}  

\IEEEoverridecommandlockouts                              

\usepackage{times}
\usepackage{graphicx}
\usepackage{xcolor}
\usepackage{algorithm}
\usepackage{algorithmicx}
\usepackage[noend]{algpseudocode}
\usepackage{amsmath}
\usepackage{booktabs}
\usepackage{graphicx}
\usepackage{amssymb} 
\usepackage{subfigure}
\usepackage{multicol}
\usepackage{hyperref}
\usepackage{bbm}
\usepackage{wrapfig}
\usepackage{arydshln}
\usepackage[font=small,labelfont=bf]{caption}
\usepackage{balance}

\DeclareMathOperator*{\argmax}{arg\,max}

\renewcommand{\baselinestretch}{0.97}

\title{\LARGE \bf
Efficient Recovery Learning using Model Predictive Meta-Reasoning
}

\author{Shivam Vats \and  Maxim Likhachev \and Oliver Kroemer 
\thanks{The Robotics Institute, Carnegie Mellon University \phantom{aaaaaaaaaa}
\tt \small \{svats, mlikhach, okroemer\} @andrew.cmu.edu
}
}

\begin{document}

\maketitle
\thispagestyle{empty}
\pagestyle{empty}

\begin{abstract}
Operating under real world conditions is challenging due to the possibility of a wide range of failures induced by execution errors and state uncertainty.
In relatively benign settings, such failures can be overcome by retrying or executing one of a small number of hand-engineered recovery strategies.
By contrast, contact-rich sequential manipulation tasks, like opening doors and assembling furniture, are not amenable to exhaustive hand-engineering.
To address this issue, we present a general approach for robustifying manipulation strategies in a sample-efficient manner.
Our approach incrementally improves robustness by first discovering the failure modes of the current strategy via exploration in simulation and then learning additional recovery skills to handle these failures.
To ensure efficient learning, we propose an online algorithm called Meta-Reasoning for Skill Learning (MetaReSkill) that
monitors the progress of all recovery policies during training and allocates training resources to recoveries
that are likely to improve the task performance the most.
We use our approach to learn recovery skills for door-opening and evaluate them both in simulation and on a real robot with little fine-tuning.
Compared to open-loop execution, our experiments show that even a limited amount of recovery learning improves task success substantially from 71\% to 92.4\% in simulation and from 75\% to 90\% on a real robot.
\end{abstract}

\section{Introduction}
It is common for robots to make mistakes while attempting a task due to noise in state estimation and actuation.
For example, a robot may miss the handle or drop the key while trying to open a door due to an incorrect handle pose estimate.
In practice, such mistakes are often handled with hand-engineered or heuristic behaviors and state machines.
While practical for relatively simple tasks in controlled environments, this approach cannot scale to systems deployed in the real-world which can fail in a variety of different ways.
Hence, there is a need for an algorithmic way to (1) discover potential failures and (2) quickly improve the robot when new failures are discovered.

\begin{figure}[ht]
\centering
\includegraphics[width=1.0\columnwidth]{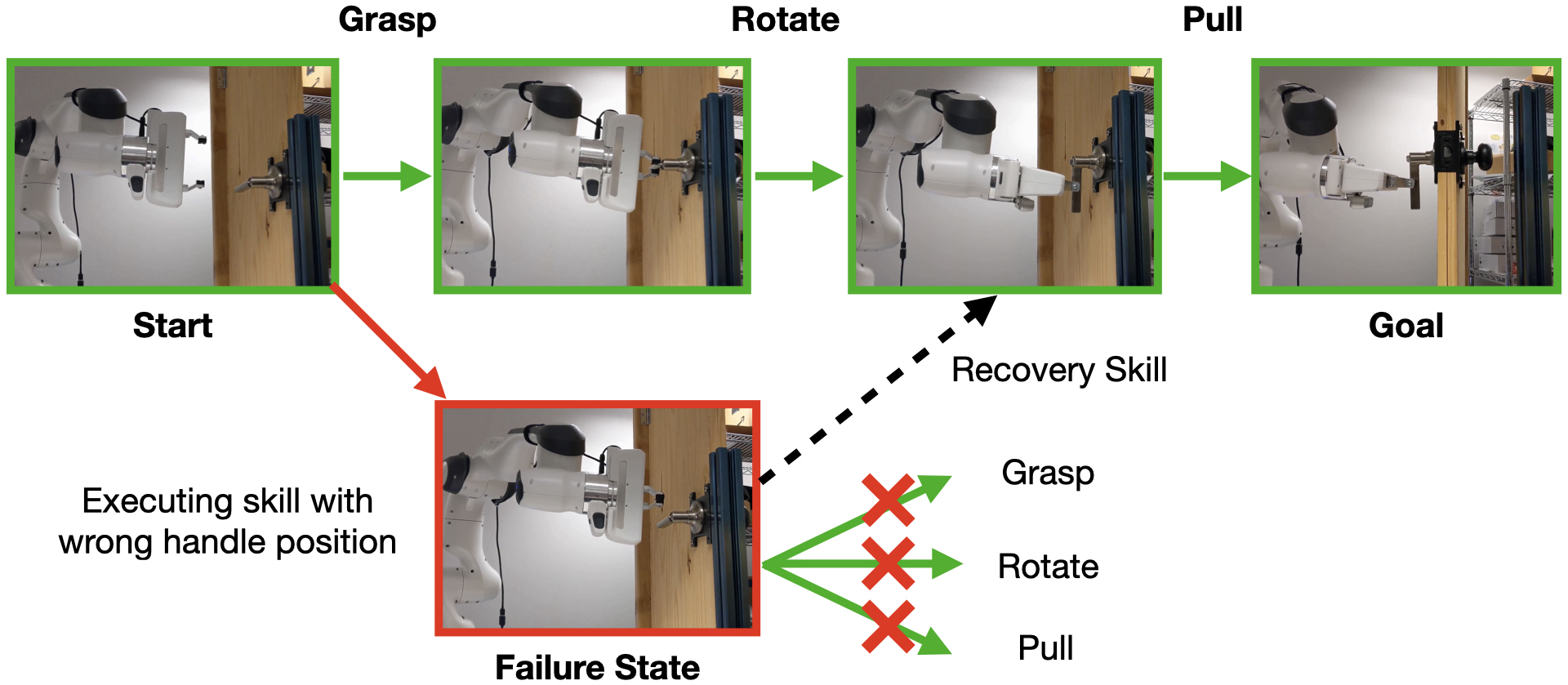}
\caption{
To open a door, the robot has to (1) grasp the handle (2) rotate it and (3) pull the door.
However, it can fail during any of these three stages due to incorrect state information and erroneously enter a failure state from which none of its skills can be applied. 
We propose an approach that (1) discovers such failures in simulation and (2) uses meta-reasoning to efficiently learn recoveries to the preconditions of the robot's existing skills.
}
\label{fig:teaser}
\end{figure}

To this end, we propose an approach to incrementally improve a robot's robustness by discovering potential failures in simulation and learning \emph{recovery skills} that allow the robot to recover. 
While our approach can robustify against failures due to uncertainty in both execution and state estimation, we focus on the latter as it is more challenging.
We assume the robot is given a nominal set of policies which can complete the task under ideal conditions.
These could be hand-designed controllers or policies learned from human demonstrations.
In reality, the state is rarely known perfectly but is estimated using an online state estimation module.
Consequently, the robot may make mistakes during execution and enter a state from which it cannot continue--- a \emph{failure state}.
We discover such failures in simulation by executing the nominal policies under a simulated state estimation model.
Next, we cluster similar failures and learn recovery skills for each of the clusters that allow the robot to recover to the \emph{precondition} of one of its nominal policies.

There are multiple potential recoveries from every failure cluster, each corresponding to a precondition the robot could recover to.
For $n$ failure clusters and $m$ preconditions, this results in a total of $n \times m$ potential recovery skills.
Since attempting to learn all of these recoveries is computationally expensive and redundant, prior works use heuristics to choose where to recover.
Recoveries generated in such a way can be sub-optimal as these heuristics don't reason about the quality of the recovery.
For example, a common heuristic is to recover to a previous state upon detecting a failure.
However, it is preferable to recover closer to the goal in terms of execution cost.
On the other hand, it is not known \emph{a priori} if recovering close to the goal is feasible. 
To this end, we propose
\textbf{Meta}-\textbf{Re}asoning for \textbf{Skil}l \textbf{L}earning (MetaReSkill),
an algorithm that builds a predictive model of improvement in skill performance and decides online which skills to devote training resources to such that the overall task performance improves maximally.

Our main contributions are (1) a hierarchical reinforcement learning-based framework for learning recovery skills to handle failures due to state uncertainty.
Compared with open-loop execution, this improves success at opening doors from 71\% to 92.4\% in simulation and from 75\% to 90\% on a real robot.
(2) a novel meta-reasoning algorithm MetaReSkill for sample-efficient recovery learning. 
Not only does MetaReSkill improve performance significantly faster than round-robin in all our experiments, but it also achieved the best performance of round-robin using only \textbf{70\%} of the training budget in $3/5$ of the experiments. 
Additional details, experiments and videos are available at ~\url{https://sites.google.com/view/recoverylearning/home}.




\section{Related Work}
\textbf{Recovery from Failures:} Robotic systems are usually deployed with hand-designed recovery behaviors in the anticipation of failures.
Common recovery strategies include retrying the previous step~\cite{ebert2018robustness,matsuoka1999dexterous}, backtracking~\cite{wang2019robust} and hand-designed corrective actions~\cite{hsiao2010contact,sundaresan2021untangling}.
To execute a recovery, it is important to first detect~\cite{rodriguez2010failure,park2016multimodal,Luo2021autonomy,zachares2021interpreting} what kind of failure has happened or is about to happen.
Pastor et al.~\cite{pastor2012towards} propose Associative Skill Memories which associate stereotypical sensory events with robot movements.
Parashar et al.~\cite{parashar2021meta} propose an architecture for robot assembly which uses meta-reasoning to identify the cause of a failure and repair the knowledge that caused the failure.
In all these works, the recovery behaviors are either manually designed, which limits their scalability, or are generated using a heuristic, which limits their complexity and quality.
By contrast, we learn recovery behaviors with reinforcement learning which offers the possibility of learning complex recoveries.
Pacheck et al.~\cite{pacheck2019automatic} encode the robot's capabilities in Linear Temporal Logic which allows them to suggest additional skills that would make an infeasible task feasible.
However, they do not deal with failures due to state uncertainty.
More recently, there has been interest in learning a policy to recover to a safe state~\cite{thananjeyan2021recovery,wilcox2022ls3}.
However, these works learn a single recovery policy, assume full observability and do not focus on the efficiency of learning.

\textbf{Meta-Reasoning:} In cognitive science, meta-reasoning~\cite{ackerman2017meta} refers to processes that  monitor the progress of human reasoning activities (\emph{metacognitive monitoring}) and allocate time and effort devoted to cognitive tasks (\emph{metacognitive control}). 
It is a key component of human intelligence which allows the human mind to solve a wide range of problems using a fixed amount of computation and limited experience~\cite{russell1991principles,Cox2011MetareasoningT,Griffiths2019DoingMW}.
Meta-reasoning has been used in artificial intelligence to design agents that operate in resource constrained environments.
One such approach is to compute the value of computation~\cite{russell1991principles} for every computation that could be executed and pursue the computation with the highest value.


\textbf{Hierarchical Reinforcement Learning:}
Hierarchical reinforcement learning (HRL)~\cite{pateria2021hierarchical} is a powerful approach for difficult long-horizon decision making problems.
It enables autonomous decomposition of the problem into tractable sub-tasks, often building hierarchies of states and policies.
A number of prior works propose algorithms for skill discovery and learning~\cite{konidaris2009skill,Konidaris2012,bagaria2019option} with the goal of covering the whole state space
By contrast, we seek to discover and cover the part of the state space  most relevant to the task.
Second, we focus on resource efficiency due to the difficulty and high complexity of skill learning in robotics.

\section{Background}


\textbf{Options Framework:}
We model each robot skill as an option as per the options framework~\cite{sutton1999between,konidaris2018skills}.
Each option consists of three components:  (a) a robot \emph{control policy} $\pi$ (b) an \emph{initiation set} which defines the states from which the option can be executed and (c) a \emph{termination condition}  which defines the states in which the option must terminate.
In continuous spaces, the initiation set is typically estimated using a binary precondition classifier, called the precondition~\cite{konidaris2015symbol}.
The skill precondition $\rho(s): \mathcal{S} \rightarrow [0, 1]$ is a function that returns the probability that the skill can be successfully executed at a given state.


\textbf{Recovery Skill:}
A recovery skill is an option that brings the system to a state from which one of its nominal skills can be executed.
Formally, let $\Pi = \{\pi_1,\cdots,\pi_k\}$ be the set of nominal skills and let $\{\rho_1, \cdots, \rho_k\}$ be their preconditions.
We say that the robot has reached a failure state if none of the preconditions is satisfied.
A recovery skill (figure \ref{fig:teaser}) drives the robot to a safe state where at least one of the preconditions is satisfied so that the robot can complete the task.

\textbf{Acting under Uncertainty:}
The problem of acting under partial or uncertain state information is optimally solved by formulating it as a Partially Observable Markov Decision Process (POMDP)~\cite{aastrom1965optimal,kaelbling1998planning}.
A POMDP is defined by the tuple $\langle S, A, T, R, \Omega, O, \gamma \rangle$, where $S$ is the underlying state space and $\Omega$ is the observation space.
In this formulation, the robot acts on its state belief $b$, which is a probability distribution over all possible current states.
However, solving a POMDP exactly is intractable in manipulation~\cite{kaelbling2013integrated}.
Instead, we use a common heuristic technique of using a state estimator to maintain a belief over world states based on observations and actions, while the robot acts on the most likely state~\cite{spaan2012partially}.
\section{Problem Setup}

We are interested in solving a manipulation task defined by a distribution of start states $\mathcal{D}$ and a goal indicator function $f_{goal}: \mathcal{S} \rightarrow \{0, 1\}$.
The robot incurs costs based on its actions and a penalty of $c_{fail}$ if it ends up in a dead-end or is unable to complete the task in T timesteps.
We are given a set of nominal control policies \{$\pi_1,\cdots,\pi_k\}$ and a high level policy $\Pi$ which chooses among these control policies.
We assume that they can reliably complete the task under no uncertainty.
Our goal is to improve the robustness of the robot by discovering and learning additional recovery skills that can handle failures due to state uncertainty.
Formally, we seek to maximize the expected return of the high-level policy on the task distribution:
\begin{align}
\mathbb{E}_{\tau \sim \mathcal{D}}\sum_{\substack{t=0, s_0=\tau \\ a_t \sim \pi(s_t), \pi \sim \Pi(s_t)}}^T R(s_t, a_t)
\label{eq:objective}
\end{align}
where $R(s, a) = -c(s, a) - c_{fail}1_{failure}$ and  $s_t$ is the most likely state at time $t$ as determined by a state estimator.

\subsection{Hierarchical Reinforcement Learning Framework}

Instead of reasoning with low-level ground states, which are high dimensional, we build a compact symbolic skill graph $G = (V, E)$.
Each vertex in this graph is a symbolic state corresponding to a set of continuous states defined by its precondition $\rho: \mathcal{S} \rightarrow \{0, 1\}$.
There exists an edge between two vertices $u, v \in V$ if there is a skill whose precondition contains $u$ and its effect is contained in $v$.
We initialize this graph as a chain with vertices $V = \{\rho_1,\cdots,\rho_k, \rho_{goal}\}$ corresponding to the preconditions of the nominal skills and edges $E$ corresponding to the nominal policies.
Let $\pi_{ij}$ be the skill from $\rho_i$ to $\rho_j$ and $q_{ij}$ be its probability of success.
If $\pi_{ij}$ fails then we assume it ends up in an absorbing failure state $Fail$ incurring a penalty of $c_{fail}$.
$\pi_{ij}$ can be learnt using off-the-shelf RL algorithms where $\rho_i$ is the initial state distribution and $\rho_j$ is the goal condition.
While we could use $\rho_j$ to define a binary reward function for RL, this is usually impractical for high-dimensional domains.
Fortunately, $\rho_j$ can also be used to define a dense reward, for example, by computing the distance to the decision boundary or using the probability $\rho_j(s)$ as the reward.
Finally, the high level policy $\Pi$ for choosing which skill to execute at every symbolic state can be computed using Value Iteration as the skill graph is discrete.

\textbf{Precondition Chaining:} The preconditions of the nominal skills can either be hand-designed or learnt.
We use an approach similar to skill chaining~\cite{konidaris2009skill, Konidaris2012} to learn the preconditions backwards from the goal.
This involves two steps:
\begin{enumerate}
    \item We collect successful trajectories by executing the nominal policies in simulation.
    Let $\{s_1,\cdots,s_k,s_{goal}\}$ be one such trajectory consisting of only the start and end states of each policy and $s_{goal}$ is a state that satisfies the goal function.
    For every policy $\pi_i$, we learn a corresponding positive distribution $\mathcal{D}^+_i$ over its start states $S^+_i$.
    \item We train the precondition classifiers backwards from the goal.
    To learn the precondition $\rho_i$, we sample states in the vicinity of $\mathcal{D}^+_i$ and execute $\pi_i$ from there.
    We verify its success using $\rho_{i+1}$ ($\rho_{goal}$ for $\rho_k$).
    This helps us gather informative negative samples $S_i^-$ and additional positive samples which are crucial for learning a tight decision boundary.
    The precondition classifier $\rho_i$ is trained using $S_i^+$ and $S_i^-$.
\end{enumerate}

\section{Approach}
Our approach consists of two steps- failure discovery and learning recovery skills using meta-reasoning.

\subsection{Failure Discovery}
 \begin{figure}[t]
 \centering
\includegraphics[width=1.0\columnwidth]{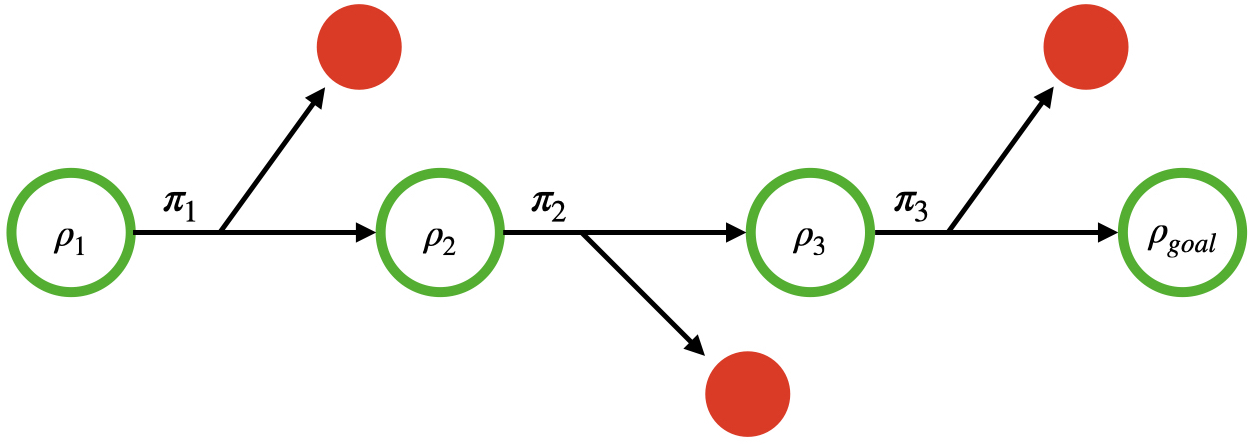}
\label{fig:failure_discovery}
 \caption{\textbf{Failure Discovery:} We execute the nominal skills under a simulated state estimator to induce failures (shown in red).
 These failure states $\in S$ are clustered into failure modes using a Gaussian Mixture Model (GMM).
 This GMM is used as a failure classifier during execution.
}
\end{figure}

We procedurally generate failure states in simulation by executing the nominal skills under noisy state information.
Concretely, let $s$ be the true current state and $o$ be a noisy observation.
While the true state is known to us in simulation, we provide only the observation to the nominal skills.
Because of the mismatch between $o$ and $s$, the skill may not work as intended and the robot may end up in a new state $s'$ with observation $o'$.
If none of the existing skills is applicable at $s'$, we record $s'$ as a failure state as the robot would not be able to recover from it even if it could observe the true state.
Note that we do not record $o'$ as it may not even be a valid world state.
While a recovery for $s'$ does not allow the robot to deal with its current observation $o'$, it will be useful when the robot observes $o \approx s'$.
We propose two failure discovery strategies:
\begin{enumerate}
\item \textbf{Pessimistic Discovery:} The robot executes its nominal policies open-loop under simulated high state uncertainty.
This strategy discovers a larger and more diverse set of failures than what may actually be encountered during execution.
While this makes recovery learning computationally more expensive, it doesn't require a model of the state estimator.
\item \textbf{Early Termination:} The robot executes its nominal policies using observations from a simulated state estimator and terminates if none of the preconditions are satisfied.
This strategy discovers a more accurate failure distribution and is preferable if a model of the state estimator is available.
\end{enumerate}

Let $S_{fail}$ be the set of failures discovered.
We cluster $S_{fail}$ into $n$ failure clusters $\{\rho_1^f,\cdots,\rho_n^f\}$
and add them as states to our symbolic skill graph.
For failures discovered using the early termination strategy, the size of a failure cluster corresponds to the likelihood that the robot will end up in that failure.
Both of these failure discovery strategies lead to recoveries that provide significant improvement in performance over heuristic recovery strategies in our experiments.




\subsection{Meta-Reasoning for Skill Learning (MetaReSkill)}

The robot may recover to one of the $k+1$ preconditions $\{\rho_1,\cdots,\rho_k, \rho_{goal}\}$ from every failure cluster.
However, many of these recoveries are redundant or infeasible.
Instead of trying to learn all of them, our algorithm identifies and prioritizes the most promising recoveries.

\begin{figure}[t]
\centering
\includegraphics[width=1.0\columnwidth]{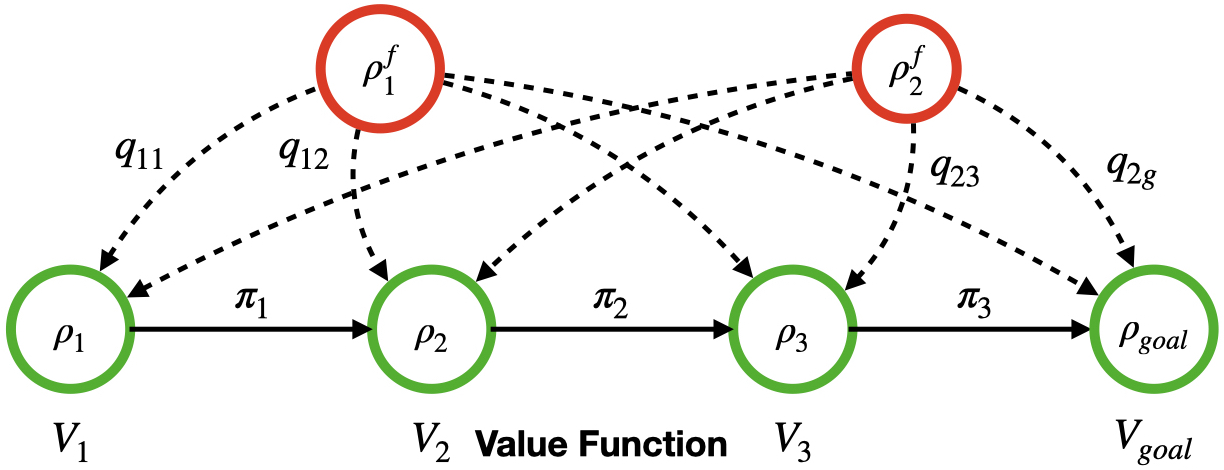}
\caption{\textbf{Optimistic Recovery Learning:} We learn recoveries using the most likely state heuristic, i.e., we optimistically assume the state becomes fully observable after the robot ends up in a failure state. 
We can potentially learn a recovery (shown in dotted lines) to each of the preconditions.
At the start of recovery learning, none of these recoveries have been learnt and the robot always incurs a penalty $c_{fail}$ for failing.
After some training, the recoveries are partially learnt and have success probabilities $q_{ij}$.
Note that it is preferable to recover to preconditions closer to the goal as they have a higher value than those away from the goal.
}
\label{fig:recovery_learning}
\end{figure}

\textbf{Value of Failures (VoF):}
We define the Value of Failures to measure the performance of the robot at its failure states.
Consider a failure cluster $\rho_i^f$ in figure~\ref{fig:recovery_learning} at the start of recovery learning.
$\rho_i^f$ is not connected to any precondition as the success probability $q_{ij}$ of all the recoveries is 0.
Hence, the value of this failure cluster is $V(\rho_i^f) = -c_{fail}$.
With further training, the value improves to 
$$\max_{j} q_{ij}V(\rho_j) - (1 - q_{ij})c_{fail}$$
$\rho_j$'s are high value states as nominal skills can be executed reliably from there.
To take into account multiple failure modes, we take a weighted sum of the values of all the failure clusters:
\begin{align}
VoF = \sum_{i}\frac{|\rho_i^f|}{\sum_j{|\rho_j^f|}}V(\rho_i^f)
\label{eq:failure_value}
\end{align}
where $|\rho_i^f|$ is the number of ground states in the cluster $\rho_i^f$.

\begin{figure}[t]
\centering
    \includegraphics[width=0.45\columnwidth]{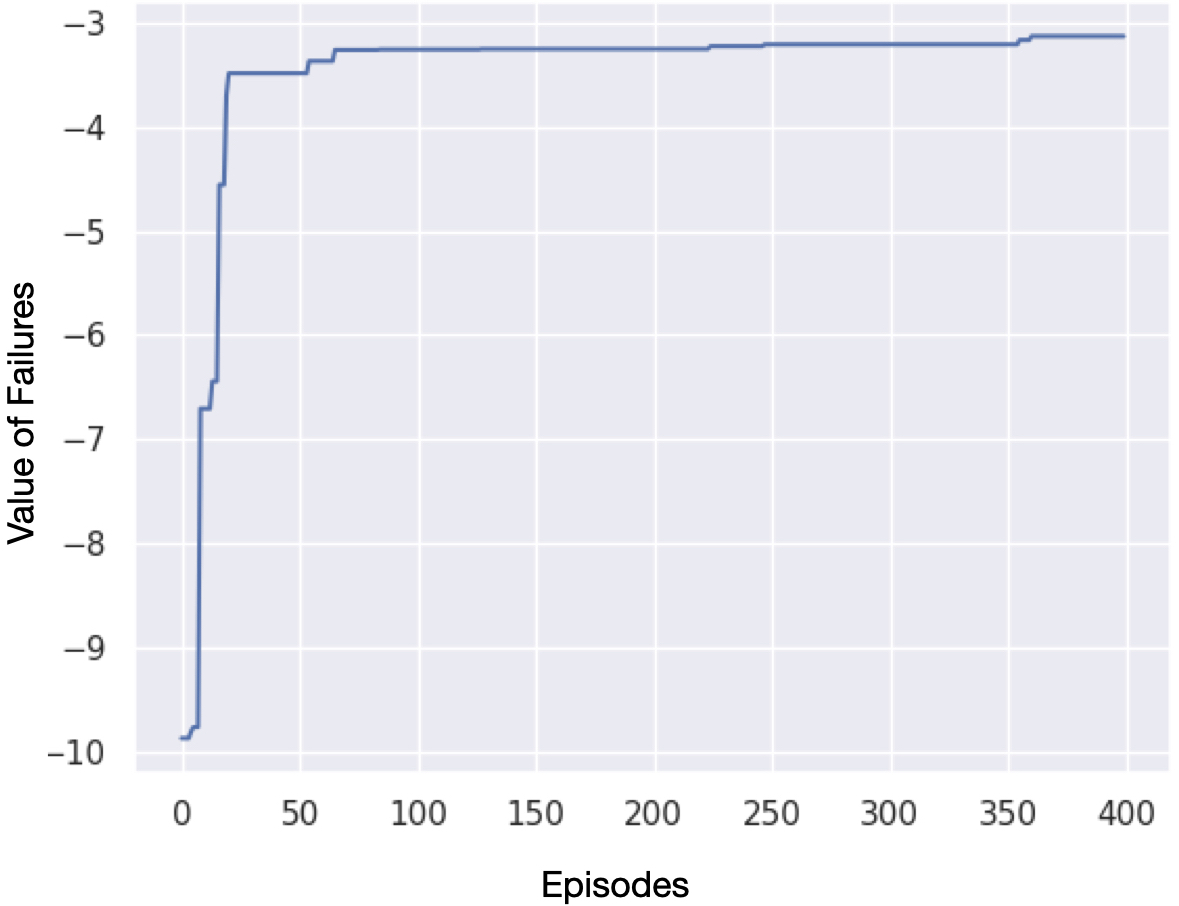}
    \includegraphics[width=0.50\columnwidth]{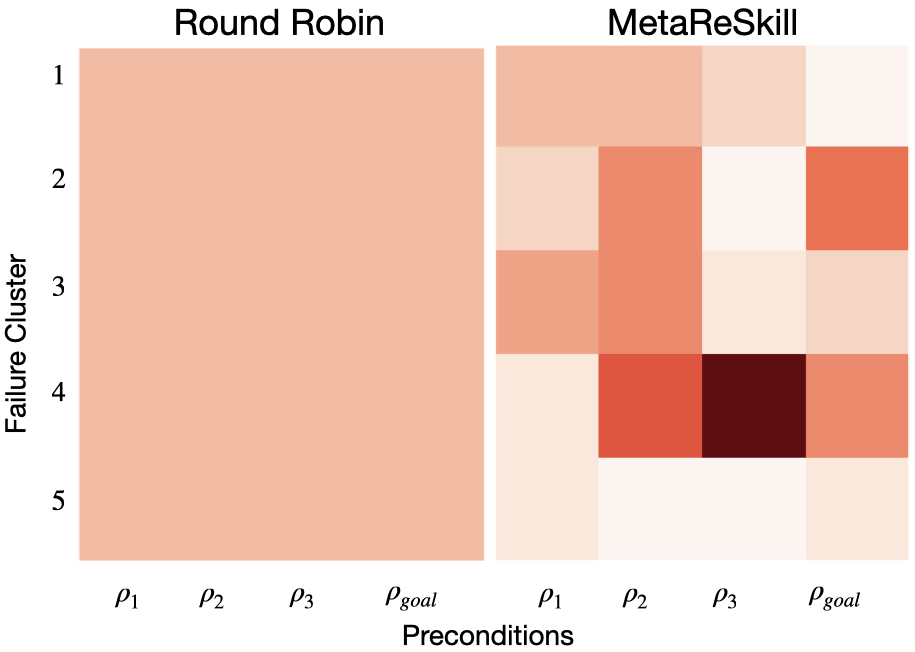}
  \caption{\emph{(left)} \textbf{VoF Saturates:} After quick initial improvement, the objective saturates when recoveries are trained in a round-robin order.
  \emph{(right)} \textbf{Allocation:} We show how many rounds each of the 20 recoveries were trained for by Round-robin and MetaReSkill.
  Each recovery is identified by the pair $(i, j)$, where, $i$ is the failure cluster it is meant for and $j$ is the precondition it recovers to.
  Round-robin trains all of them equally while MetaReSkill prioritizes a small number of promising recoveries that improve the VoF by the most.
  }
  \label{fig:allocation}
\end{figure}

Intuitively, a high VoF implies that failures during execution are less problematic as the robot is highly confident of recovering.
Learning recoveries that optimize the VoF improves robustness to failures.
A good first meta-strategy is to train all the recoveries in a round-robin manner as we do not know \emph{a priori} what the best recovery from every failure mode will be.
While this works well in the initial stages of learning, we observed in our experiments that it is quite inefficient as the VoF quickly saturates (figure~\ref{fig:allocation}).
To address this issue, we propose a meta-reasoning algorithm that tracks the progress of all the recoveries and chooses which failure modes to focus on and which precondition to recover to such that the VoF improves maximally with high probability in every training episode.

\begin{algorithm}
\centering
    \begin{algorithmic}[1]
        \Procedure{train}{}
        \State{$Q_{ij} \leftarrow$ queue of max size $w, \forall i,j$}
        \For{$1\leq t \leq B$}
            \If{all policies have been trained $\geq K$ times}
            \For{all $i, j$}
                \State{$T \leftarrow$ transition matrix of skill graph}
                \State{$T(i, j) \leftarrow q_{ij}^U$}
                \State{$J^U_{ij} \leftarrow \textsc{objective}(T, R)$}
            \EndFor
            \State{$(i^*, j^*) \leftarrow \argmax$ $J^U$}
            \Else
                \State{$(i^*, j^*) \leftarrow$ least trained policy}
            \EndIf
            \State{train recovery $(i^*, j^*)$ for $\eta$ episodes}
            \State{$q \leftarrow$ estimate new success rate of $\pi_{i^*j^*}$}
            \State{$q_{best} \leftarrow \max(q, q_{i^*j^*})$}
            \State{$q_{i^*j^*} \leftarrow q_{best}$, $Q_{ij}$.insert($q_{best}$)}
            \State{$q^U_{i^*j^*} \leftarrow$ \textsc{ComputeUCL}($i^*,j^*$)}
        \EndFor
        \Return{$\pi_{ij}, \forall i,j$}
        \EndProcedure
        
        \Procedure{Objective}{$T, R$}
            \State{$V \leftarrow$ \textsc{ValueIteration}($T, R$)}
            \State{}
            \Return{$\sum_{i}\frac{|\rho_i^f|}{\sum_j{|\rho_j^f|}}V(\rho_i^f)$} \Comment{Value of Failures}
        \EndProcedure
        
        \Procedure{ComputeUCL}{$i, j$}
            \State{$\Delta Q_{ij} \leftarrow$ compute forward differences of $Q_{ij}$}
            \State{$n \leftarrow |\Delta Q_{ij}|$}
            \State{$\Delta q^U_{ij} \leftarrow \overline{\Delta Q_{ij}} + t_{(1-\alpha)/2,n-1}\frac{s}{\sqrt{n}}$ \Comment{UCL of ROI}}
            \State{$q^U_{ij} \leftarrow q_{ij} + \Delta q^U_{ij}$}
            \Return{$q^U_{ij}$}
        \EndProcedure
    \end{algorithmic}
    \caption{Meta-Reasoning for Skill Learning}
    \label{alg:value_uci}
\end{algorithm}

\textbf{Model of Task Performance:}
The key idea of MetaReSkill is to build a model of improvement in task performance by estimating the \emph{rate of improvement} (ROI) of individual skills.
We use confidence interval estimation to compute optimistic upper bounds $\Delta q_{ij}^U$ of the ROI $\Delta q_{ij}$ of the success probabilities $q_{ij}$ of all the recoveries.
This provides an optimistic estimate of how much a recovery could improve after another round of training.

Let $\theta$ be a parameter we wish to estimate.
An $\alpha$-confidence interval~\cite{hines2008probability} for $\theta$ is an interval $(L, U)$ such that $\theta$ is contained in the interval with confidence $\alpha$.
This also implies that $U$ is an upper bound on $\theta$ at least with confidence $\alpha$.
Let $\theta_1,\cdots,\theta_w$ be a random sample of the parameter.
Under the assumption that the underlying population is normally distributed, the mean $\mu$ of the distribution lies in the following interval with probability $\alpha$:
$$\overline{\theta} - t_{(1-\alpha)/2,w-1}\frac{s}{\sqrt{w}} \leq \mu \leq \overline{\theta} + t_{(1-\alpha)/2,w-1} \frac{s}{\sqrt{w}}$$
where $t_{\alpha,w}$ is the upper $\alpha$ percentage point of the student's $t$-distribution with $w$ degrees of freedom and $s$ is the standard error.

We compute the upper confidence limit (UCL) of $\Delta q_{ij}$ using only the $w$ most recent forward differences of $q_{ij}$ as the rate of improvement is a non-stationary quantity ($w$ is a domain-dependent hyper-parameter).
For every recovery with current success probability $q_{ij}$, $q_{ij}^U = q_{ij} + \Delta q^U_{ij}$ is an optimistic upper bound on its success probability after an additional round of  training.
Let $J^U_{ij}$ be the VoF computed by replacing $q_{ij}$ with $q_{ij}^U$ in the transition matrix of the recovery learning graph~\ref{fig:recovery_learning}.
$J^U_{ij}$ is then an optimistic prediction of the VoF if we were to train $\pi_{ij}$ for another round.
Our algorithm greedily picks a recovery for training that promises the highest VoF in the next round.
We initialize MetaReSkill with $K$ rounds of round-robin to estimate the UCL.
Priors on $\Delta q_{ij}$, if available, can further speed up learning.

\section{Experiments} \label{experiments} 
We evaluate our approach on the task of door opening under noisy handle position information  both in simulation and in the real world.
The goal is to open a door by at least $0.3$ rad with the Franka Panda robot under high initial state uncertainty.
%


\begin{figure*}[th]
  \begin{center}
    \includegraphics[width=1.0\textwidth]{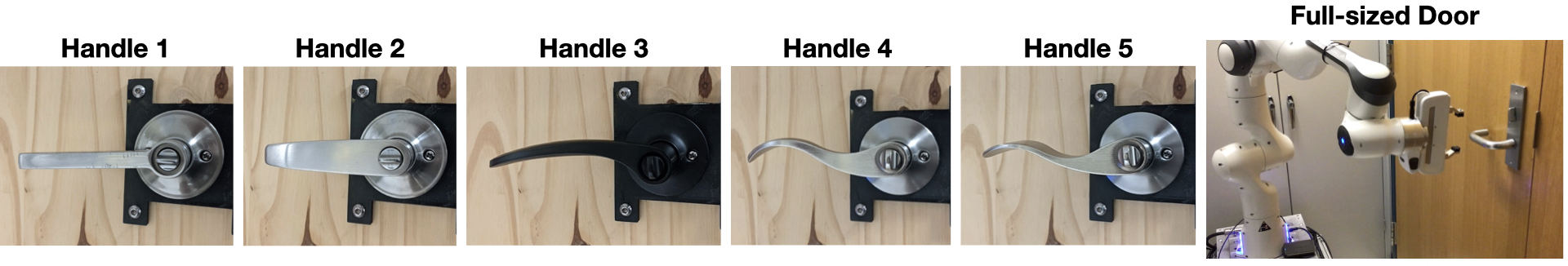}
  \end{center}
  \caption{\textbf{Handles Used in Evaluation:} We compare our approach with open-loop execution on 5 different lever latch handles and a full-sized door with a real robot.
  Only handle 1 and a small door was used during training; handles 2-5 and the full-sized door are unseen.
  }
  \label{fig:handles}
\end{figure*}

\textbf{Simulation Environment:}
We adapt the door environment from the MuJoCo-based robosuite~\cite{todorov2012mujoco,robosuite2020} framework to match our real door.
The world state is 18 dimensional and includes the robot's joint angles and poses of the door and the handle.
The initial state uncertainty is sampled from $\mathcal{N}(0, \sigma=2cm)$ each in the x, y and z positions of the handle.
We design 3 open-loop skills to be executed in sequence - \textsc{ReachAndGraspHandle}, \textsc{RotateHandle} and \textsc{PullHandle}- as the nominal skills. Each skill consists of one or more 7D waypoints that the robot tries to reach using task space impedance control, where each target consists of a gripper open/close state and a 6D end-effector pose.
These skills are able reliably to open doors in simulation and the real world if accurate state information is available.

\begin{table}
\centering
    \begin{tabular}{lccc}\toprule
                             & Success Rate (\%) &     Cost (m)   \\ \midrule
                             
        Recovery-skills (Ours)    &  \textbf{92.4}    & 0.95 ($\pm$ 0.34) \\
        Retry              & 66.9     & 0.80 ($\pm$ 0.02)\\
        Recover-to-prev    & 75.5     & 0.86 ($\pm$ 0.19)\\
        Recover-to-start   & 73.6     & 0.87 ($\pm$ 0.26)\\
        No-recovery        & 64.4     & 0.80 ($\pm$ 0.02)\\
        Open-loop          & 71.0       & 0.80 ($\pm$ 0.02)\\
      \end{tabular}
    \caption{\textbf{Simulation results:} We compare recovery skills trained with our approach using  150 REPS queries with heuristic recovery strategies.
    Our approach significantly improves the success rate.
    The statistics are averaged over 5 sets of recovery skills learnt with different seeds, each evaluated 200 times using a simulated state estimator and a limit of 10 skills per evaluation.
    }
\label{table:results1}
\end{table}

\textbf{Symbolic Skill Graph:} 
We train the preconditions of the nominal skills by precondition chaining using a total of 1223 positive and negative samples.
Each precondition is a  generative classifier with the positive distribution $\mathcal{D}^+$ learnt as a Gaussian distribution and the negative distribution $\mathcal{D}^-$ learnt as a Gaussian Mixture Model.
The 3 nominal skills result in 4 symbols for the start, subgoals, and the goal.

\textbf{Recovery Skill:} Each recovery skill $\pi_{ij}$ is a parameterized skill~\cite{da2012learning} that uses a regression model to predict robot actions $\theta \in \Theta$ based on the start state $s$.
We use k-nearest neighbours regression to predict a 21D vector, a sequence of three 6D poses with respect to the initial end-effector pose and gripper open/close states, as robot actions.
Collecting data of the form $(s, \theta)$ for training this regression model involves sampling a start state $s$ from the failure mode $\rho_i^f$ and computing the robot action parameters $\theta$ for recovery to $\rho_j$ using Relative Entropy Policy Search (REPS)~\cite{peters2010relative}.
For learning a recovery to precondition $\rho$, REPS uses the reward function
$R(s) = 0.1 \log f_{\mathcal{D^+}}(s) + 10 \rho(s)$,
where $f_{\mathcal{D^+}}$ is the  probability density function of the corresponding $\mathcal{D^+}$.
Each REPS query takes about 2 minutes to solve on our Intel® Core™ i7-9700K CPU.

\subsection{Evaluation of Learnt Recovery Skills}

We first evaluate the effectiveness of our overall approach using the pessimistic failure discovery strategy we described earlier.
We execute the nominal skills 1000 times for failure discovery to collect a total of 1400 failure states which we group into 6 clusters using the Gaussian Mixture Model (GMM)~\cite{scikit-learn}.
Common failure modes include the robot missing the handle and the robot slipping while pulling the handle due to an improper grasp.
We learn recovery skills in simulation using a budget of just 150 REPS queries.
With 24 potential recovery skills, this means that each recovery policy can get only 6 data-points on average.

\textbf{Evaluation in Simulation:}
We simulate a state estimator by assuming that the standard deviation of the noise distribution halves after every robot action.
As we show in table~\ref{table:results1}, learnt recoveries are significantly better than heuristic recovery strategies in improving success rate.
Recovering from failures during door opening often requires the robot to (1) carefully move the handle so as not to weaken the grasp and (2) avoid collision with the environment.
Heuristic recoveries are unable to account for this and hence perform poorly.
Compared to open-loop execution, our approach substantially improves task success rate from $71\%$ to $92.4\%$.
This indicates that (1) the failures discovered using our pessimistic failure discovery do cover a number of failures encountered by the robot when using a state estimator and
(2) recoveries can be reliably learnt with a generic reward function defined using preconditions which promises better scalability than reward shaping.

\begin{table}
\centering
    \begin{tabular}{lcc}\toprule
                             & Open-loop (\%)     & Recovery (\%)   \\ \midrule
    Handle 1 (Train) & 75 (15/20) & \textbf{90} (18/20) \\
    Handle 2 (Test) & 50 (5/10) & \textbf{80} (8/10) \\
    Handle 3 (Test) & 80 (8/10) & \textbf{80} (8/10) \\
    Handle 4 (Test) & 80 (8/10) & \textbf{90} (9/10) \\
    Handle 5 (Test) & 60 (6/10) & \textbf{90} (9/10) \\
    \hline \\
    Full-sized door (Test) & 30 (3/10) & \textbf{50} (5/10) \\
      \end{tabular}
    \caption{\textbf{Success Rate on a Real Robot:} Learnt recovery skills significantly outperform open-loop execution on a real robot across 5 different handles.
    Open-loop fails almost half the time on handles 2 and 5 which are, respectively, the smallest and the thinnest of the 5 handles.
    By contrast, the learnt recoveries use a caging grasp to re-grasp the handle close to the handle's axis of rotation and are robust to these variations.
    We also test recovery skills on a full sized door where  the success rate of our approach drops to 50\% due to the slip-prone cylindrical handle of the door.
    }
\label{table:results3}
\end{table}

\textbf{Evaluation on a Real Robot:}
We transfer the preconditions, failure classifier, nominal skills and recovery skills learnt in simulation to a real Franka Panda robot and run experiments with  (a) 5 different lever latch handles on a small door and (b) a full-sized door in our building with a cylindrical handle (figure~\ref{fig:handles}).
We fine-tune only the gains of the impedance controller on the real robot by increasing their values to achieve similar tracking as in simulation.
As in simulation, the robot is controlled by a Cartesian-space impedance controller that executes each skill open-loop.
We evaluate the recovery skills learnt in simulation under an \emph{idealized state estimator}.
The ground-truth handle position is known to us but at $T = 0$, we only provide a noisy handle position estimate to the robot, where, noise $\sim \mathcal{N}(0, \sigma=2cm)$.
The robot executes its nominal skill using this noisy state information.
At $T = 1$, by the time the robot finishes executing the first skill, we assume that the state estimator has converged to the ground-truth.
Hence, the robot has access to the accurate handle position at this point.
The robot uses its preconditions to check if any of its nominal skills can be executed.
If so, it executes the remaining nominal skills.
If not, it uses the failure classifier to identify the failure mode and execute the best recovery from that mode.

We compare our approach (\textsc{recovery}) with open-loop execution (\textsc{open-loop}) of nominal skills (table~\ref{table:results3}).
The success rate of \textsc{open-loop} is sensitive to the handle and varies from $50-80\%$
By contrast, \textsc{recovery} improves the success rate to $80-90\%$ consistently across all of the 5 handles even though it was trained only for handle 1.
Both \textsc{open-loop} and \textsc{recovery} struggle at the full-sized door due to the slippery cylindrical handle.
However, our approach still does significantly better than \textsc{open-loop} which only succeeded in $3/10$ attempts.
We expect recovery performance to improve with further training and by the use of a good state estimator which will enable closed-loop behavior.
Importantly, our approach did not induce any additional failures which indicates good transfer of the preconditions and failure classifier learnt in simulation.

\subsection{Evaluation of MetaReSkill}
\begin{figure}[t]
\centering
\includegraphics[width=1.0\columnwidth]{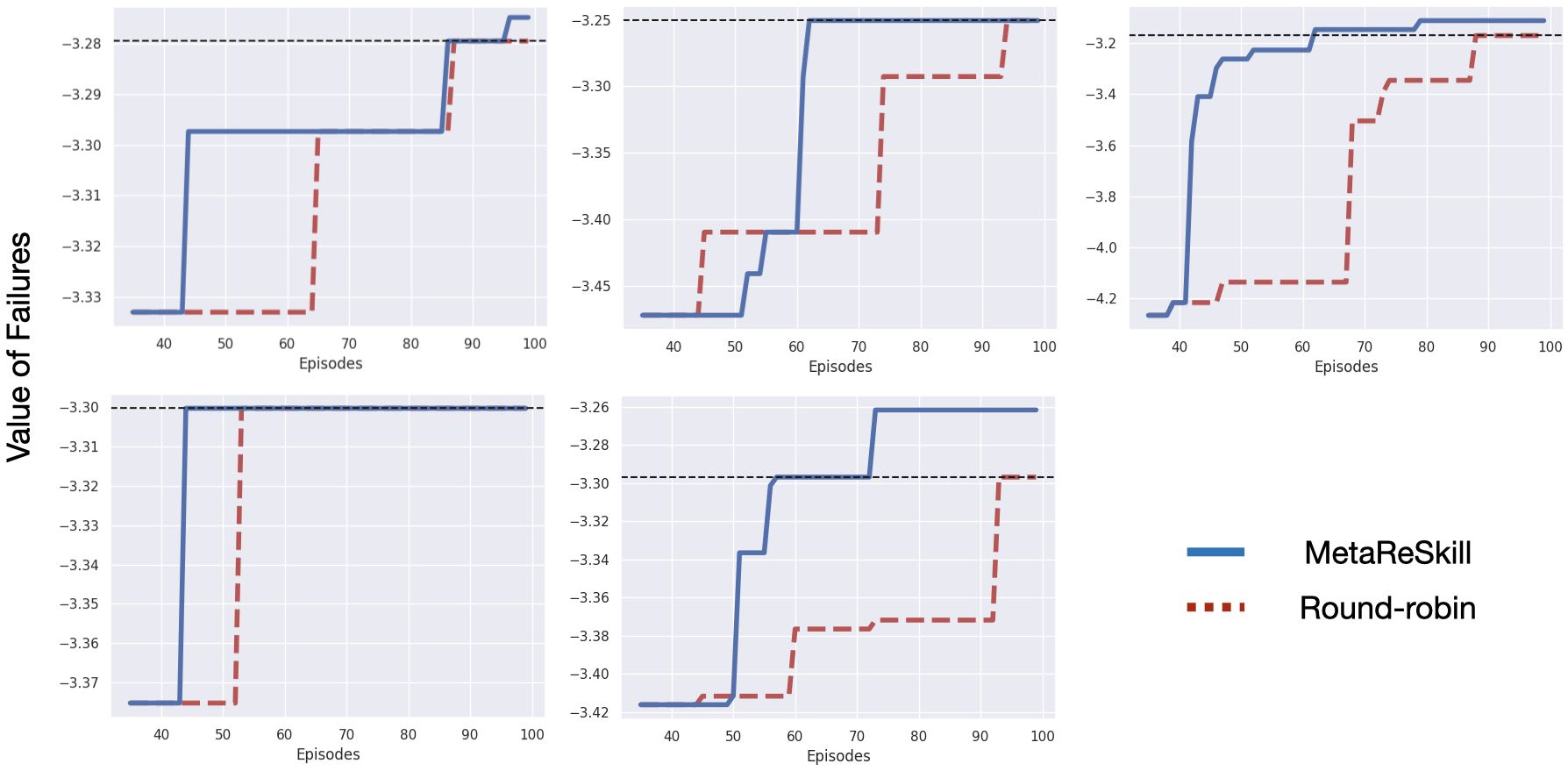}
\caption{We compare MetaReSkill with round-robin learning of recoveries over 100 REPS training episodes with 5 different seeds.
MetaReSkill is initialized by training each recovery twice initially in a round-robin manner.
Hence, we see a difference in performance from episode 40 when MetaReSkill kicks in.
MetaReSkill immediately focuses on the most promising recoveries which allows it to optimize the VoF significantly faster, also converging to a better VoF in every trial.
}
\label{fig:rr_vs_uci}
\end{figure}

In this evaluation, we assume that we have access to an accurate model of the state estimator so that we can estimate the failure distribution accurately.
We discover failures using our early termination discovery strategy along with a simulated state estimator that halves the standard deviation of the noise distribution after every robot action.
We discover a total of 2000 failure states which we group into 5 clusters using the GMM.
These failures are less diverse than the failures discovered by pessimistic discovery and are more concentrated near the door handle.
We compare round-robin learning of recoveries with MetaReSkill in figure~\ref{fig:rr_vs_uci} using 5 different seeds.
We use $\alpha = 0.95$ to compute the confidence interval, window size $w$ of 3 and query REPS for a new data-point in every round, i.e. $\eta = 1$.
We initialize the UCL estimates by training every recovery twice in a round-robin order.
Not only does MetaReSkill improve significantly faster than round-robin, but also it converges to a better objective in all the trials.
In $3/5$ trials, MetaReSkill used only \textbf{70\%} of the training budget to achieve the best objective achieved by round-robin, i.e., $1$ hour earlier.
This shows that it can make better use of training resources to improve robustness.

\section{Conclusion} 
\label{sec:conclusion}
We propose a scalable algorithmic framework to efficiently robustify a given manipulation strategy against failures due to state uncertainty.
Our method consists of discovering failures in simulation by evaluating the given strategy under simulated state uncertainty and then using meta-reasoning to efficiently train recovery skills.
Our algorithm Meta-Reasoning for Recovery Learning (MetaReSkill) monitors the progress of all potential recovery skills during training and adaptively chooses which recoveries to train to best improve the robot's robustness.
Compared to baselines, the learnt recovery skills significantly improve task success both in simulation and in the real world.
We also find that MetaReSkill makes a much better use of computation than round-robin skill learning.
In our future work, we would like to combine failure discovery in the real world with discovery in simulation to further improve robustness.
Finally, we are also interested in investigating if we can provide theoretical bounds on the performance of MetaReSkill.

\section*{Acknowledgments}
This work was in part supported by ARL grant W911NF-18-2-0218 and ONR grant N00014-18-1-2775.

\clearpage
\balance
\bibliographystyle{IEEEtran}
\bibliography{references}  


\end{document}